\documentclass[runningheads]{llncs}
\usepackage{graphicx}
\usepackage{booktabs}
\usepackage{multirow}
\begin{document}
\title{Global Prompt Cell: A Portable Control Module for Effective Prompt Tuning}
\titlerunning{Global Prompt Cell}
%

\author{Chi Liu \and
Haochun Wang \and
Nuwa Xi \and
Sendong Zhao \and
Bing Qin
}
\authorrunning{C. Liu et al.}
%
\institute{
Harbin Institute of Technology\\
\email{\{cliu, hcwang, nwxi, sdzhao, bqin\}@ir.hit.edu.cn}}
\maketitle              

\begin{abstract}

As a novel approach to tuning pre-trained models, prompt tuning involves freezing the parameters in downstream tasks while inserting trainable embeddings into inputs in the first layer. 
However, previous methods have mainly focused on the initialization of prompt embeddings. The strategy of training and utilizing prompt embeddings in a reasonable way has become a limiting factor in the effectiveness of prompt tuning. 
To address this issue, we introduce the Global Prompt Cell (GPC), a portable control module for prompt tuning that selectively preserves prompt information across all encoder layers. Our experimental results demonstrate a 5.8\% improvement on SuperGLUE datasets compared to vanilla prompt tuning.

\end{abstract}
\section{Introduction}
Prompt-based methods can be classified into two categories: discrete prompt tuning 
\cite{radford2018improving,brown2020language} and continuous prompt tuning \cite{lester2021power}. Discrete prompt tuning transforms the task into a ``fill-in-the-blank" format and then utilizes a pre-trained language model to predict the answer, which operates similarly to the masked language model (MLM) \cite{devlin2019bert}. In subsequent research, continuous prompt tuning\cite{lester2021power} introduced soft prompts (i.e., prompt embeddings) to replace manual templates, which consist of special tokens with adjustable embeddings. We refer to continuous prompt tuning as ``prompt tuning" for simplicity.

Vanilla prompt tuning, which concatenates prompt embeddings with input tokens in the first layer and updates only the parameters of the prompt embeddings during the training phase, has several limitations \cite{lester2021power,Yusheng}.
Firstly, since the effectiveness of prompt embeddings is highly related to the length, it is necessary to use prompt embeddings with hundreds of tokens in length to achieve better downstream task performance, as suggested by \cite{lester2021power} and \cite{Yusheng}. However, long prompt embeddings also reduce the possible length of input text. In addition, prompt learning requires a longer time to converge compared to full fine-tuning \cite{Su_2022}, and its effectiveness still has significant room for improvement.

These drawbacks are due to the traditional approach of inserting prompt embeddings into the input layer and concatenating them with token embeddings for model input in prompt learning. However, prompt embeddings have significant differences compared to token embeddings. Firstly, prompt embeddings have not undergone pre-training and require more optimization steps compared to token embeddings. Secondly, prompt embeddings do not have semantic information but serve as task-specific vectors to guide the model for downstream tasks. Finally, in pre-trained models with multiple layers, prompt learning freezes the parameters during training, updating only the bottom-level prompt embeddings, which can cause long-distance backpropagation to result in vanishing gradients and slow convergence\cite{deeplearing}. These reasons indicate the need to design better training and utilization methods for prompt embeddings, instead of using the same training method as token embeddings.
The RNNs contain a hidden unit that can preserve important information during sequence iterations and guide the model to output results \cite{rnns}. RNNs can be unfolded into a long sequence, and through the hidden unit, the model can selectively integrate information from different times from the beginning to the end. The input at the initial time can guide the judgment at the final time, and this approach is beneficial in alleviating the problem of vanishing gradients, reducing forgetting, and speeding up convergence.

Prompt embeddings suffer from information loss when passing through each layer of transformers. Although residual connection modules exist within each layer, these modules operate on the entire sequence, including the token embeddings, rather than the prompt embeddings alone. Therefore, due to the reasons mentioned earlier, we need to design a dynamic information fusion mechanism specifically for the part corresponding to the prompt embeddings. Inspired by RNNs, we propose the Global Prompt Cell (GPC) to address the aforementioned issues.

The Global Prompt Cell (GPC) consists of two units: the remembering unit and the forgetting unit. 
Remembering unit should be applied to the prompt embeddings before passing through the transformer layer because information loss occurs after passing through the layers. Therefore, we use remembering unit to selectively remember certain information. On the other hand, forgetting unit should be applied to the prompt embeddings after passing through the transformer layers. The forgetting unit can selectively forget some information in the latest prompt embeddings to fuse with the previous prompt embeddings.
By utilizing the same remembering unit and forgetting unit for every model layer, our approach gathers information regarding prompt embeddings from all layers, enabling it to guide the model to achieve better downstream task performance. As a result, prompt embeddings are no longer simply a vector concatenated with token embeddings, but also a control module that aids PTMs in making better decisions. Moreover, since GPC only acts on the prompt embeddings updates between layers, it can be considered a plug-in module and requires only a small number of additional parameters. Lastly, we eliminate the verbalizer in vanilla prompt tuning and use classification heads for downstream tasks, reducing the difficulty of selecting optimal verbalizers for various tasks.
To summarize, our contributions are three-fold:
    
\begin{enumerate}
\item 
We propose a new training and utilization method for prompt embeddings called the Global Prompt Cell (GPC). To our knowledge, GPC is the first method that specifically aims to improve the training and utilization of prompt embeddings, reduce forgetting during training, and ultimately enhance downstream task performance.
Experiments prove its effectiveness and show its validity in architecture with ablation study.
\item Our method simplifies prompt tuning by discarding the verbalizer, which further reduces the time and computing consumption to select the optimal verbalizer, and achieves even better results.
\item Our method can be viewed as an easy-to-implement plug-in module with only a few additional parameters, which makes GPC both model-agnostic and task-agnostic.
\end{enumerate}





\section{Related Work}


\subsection{Pre-trained Language Model}

Recently substantial works have shown that pre-trained models (PTMs) can learn universal language representations through pre-trained tasks on large corpora, which are beneficial for downstream NLP tasks and can avoid training a new model from the beginning~\cite{qiu2020pre}.




A representative application of PTMs is using encoder-based models for classification tasks~\cite{sun2019fine}. Encoder-based models are composed of multi-layer transformer encoders, which include a self-attention module and a multi-layer perceptron (MLP), and are optimized for performance through measures such as intra-layer residual connections. BERT is a classic encoder-based model commonly used for classification tasks~\cite{devlin2019bert}. In BERT, each input sentence is concatenated with a [CLS] token at the beginning. After passing through the encoder of each layer, the [CLS] token is utilized as a classification indicator to produce the final result.


\subsection{Prompt Tuning}

\begin{figure}[h] 
\centering 
\includegraphics[width=0.7\columnwidth]{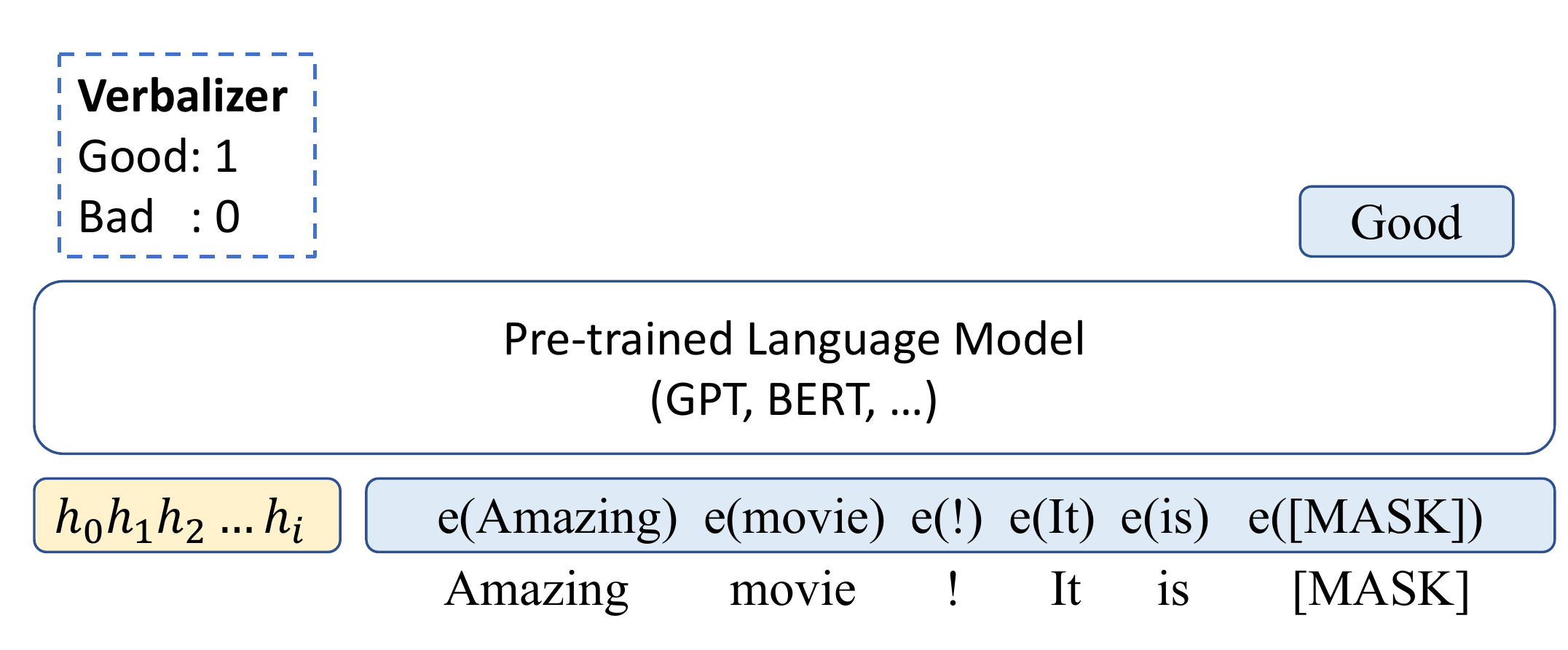} 
\caption{Prompt-tuning model.} 
\label{p-tuning} 
\end{figure}

GPT-3~\cite{brown2020language} has revolutionized downstream tasks by transforming them into generation tasks through the addition of prompt-like hints. This allows the model to generate results directly in few-shot or zero-shot learning scenarios. Prompt consists of a template for transforming the input text and a verbalizer for matching the generated words to the actual task labels. It can be designed manually by domain experts~\cite{schick-schutze-2021-exploiting,schick-schutze-2021-just} or automatically~\cite{shin-etal-2020-autoprompt,gao-etal-2021-making}, but at the expense of low explainability~\cite{shin-etal-2020-autoprompt}.

Prompt tuning methods with prompt embeddings have been explored in recent studies~\cite{hambardzumyan-etal-2021-warp,qin-eisner-2021-learning,zhong-etal-2021-factual,liu2021gpt}. Prompt embeddings are trainable embeddings rather than natural language tokens. Vanilla prompt tuning, proposed by \cite{lester2021power}, concatenates the prompt embeddings with the token embeddings in the first encoder layer.

To use prompt tuning for downstream tasks, we first encode a sequence of discrete input tokens $X = {x_0, x_1, ..., x_n}$ into embeddings $X_e = {e(x_0), e(x_1), ..., e(x_n)}$ using a pre-trained language model $M$. We then obtain the prompt embeddings $P = {p_1, p_2, ..., p_n}$, where $p_i$ is the prompt embedding for $i \in n$, and concatenate the prompt embeddings and token embeddings to form the complete input $I = {P;X_e}$. The model output embedding of the [MASK] token is then fed into a classifier to predict the target token, and the verbalizer is used to obtain the actual label.

Current research on prompt tuning architecture retains the verbalizer as a ``necessary" component, despite its high computational cost, both in terms of time and computing resources, even when constructed automatically. Despite the improvements observed in downstream tasks, the results of prompt tuning can still be inconsistent or less-than-ideal.

Prompt tuning is a newly-arising paradigm that requires significantly more training time than fine-tuning to achieve the same performance~\cite{Yusheng}, despite the two paradigms only differing in model inputs. This is due to the fact that although prompt embeddings are fundamentally different from token embeddings in terms of initialization and acquisition, vanilla prompt tuning treats prompt embeddings in a similar manner to prompt-like token vectors and processes them together using the same method.

\subsection{Model Degradation and Prompt Forgetting}

When training deep artificial neural networks, two problems become increasingly challenging for model optimization. The first problem is vanishing gradients~\cite{hochreiter1998vanishing}, which impedes convergence from the outset, but can be largely mitigated through normalized initialization and intermediate normalization layers. The second problem is the degradation problem~\cite{he2016deep}: as the network becomes deeper, accuracy saturates and then degrades rapidly, despite not being caused by overfitting, which suggests that not all systems are equally easy to optimize~\cite{he2016deep}. Skip connections~\cite{skip-connect} are a common countermeasure for model degradation, as seen in ResNet~\cite{he2016deep} and Transformer~\cite{vaswani2017attention}, where residual connections are employed. Skip connections serve as a reminder for the model to retain previous information and prevent forgetting.

Prompt tuning is also affected by these problems since it occurs at the lowest layer. No previous studies have addressed the issues of model degradation and prompt forgetting in prompt tuning, and we take these problems into consideration.

\section{Method}
 In this section, we present the architecture of our method, which incorporates the implementation of Global Prompt Cell (GPC) into an encoder-based model. Drawing inspiration from RNN models, we utilize a method to effectively store previous states of prompt embeddings in various layers, allowing the model to zoom out and concentrate on the prompt from a wider perspective.


\subsection{GPC between the Encoders}

\begin{figure*}[htbp] 
\centering 
\includegraphics[width=1
\textwidth]{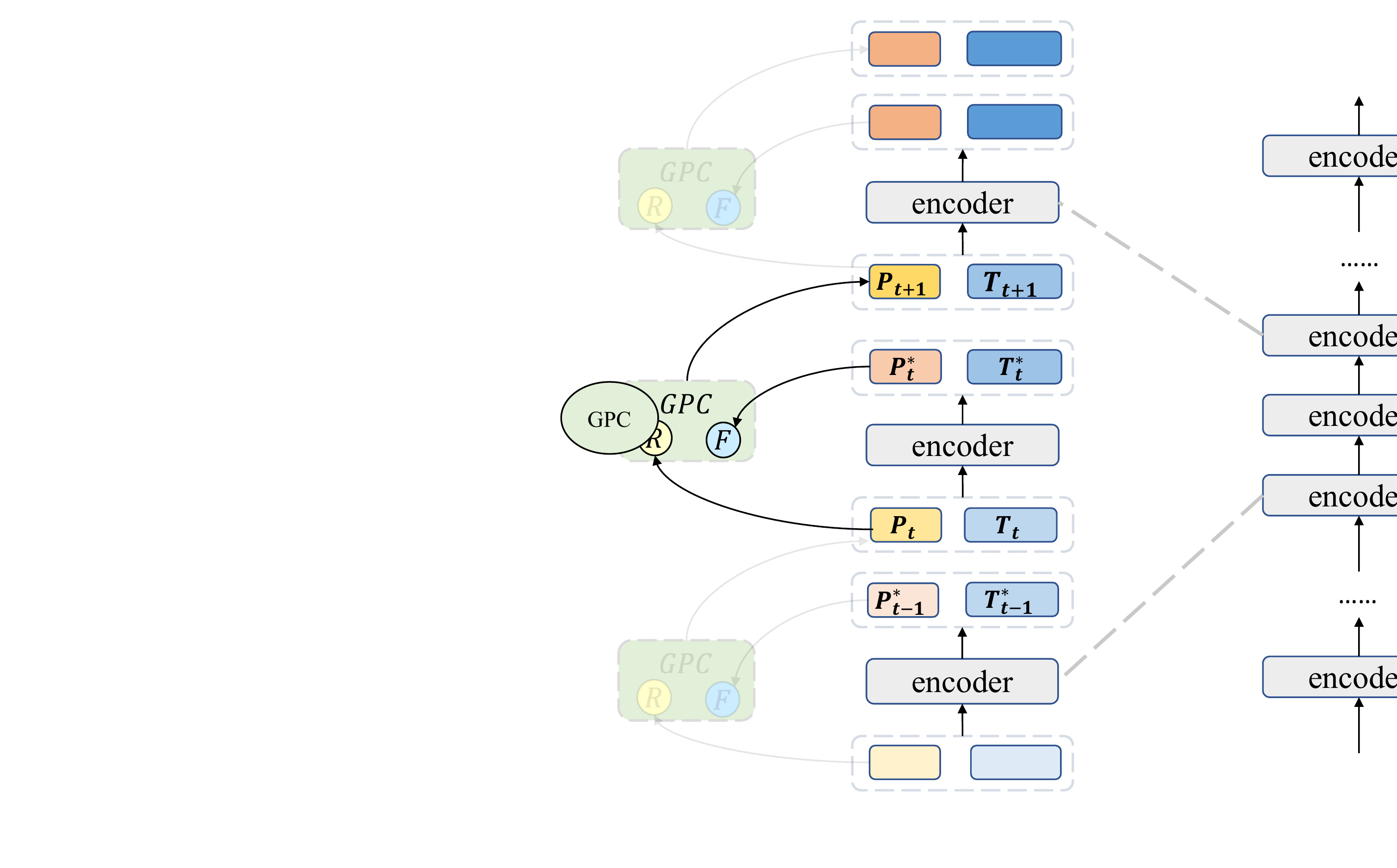} 
\caption{Global Prompt Cell Model (best viewed in color).} 
\label{gpc_model} 
\end{figure*}


In the Transformer model, residual connections are employed to mitigate the problem of degraded training accuracy.~\cite{vaswani2017attention}. Building on this idea, GPC employs a more advanced connection mechanism around each pair of adjacent encoders. Specifically, GPC operates exclusively on the prompt portion of each vector. We use $P$ to represent prompt embeddings and $T$ to represent token embeddings, with the two concatenated to form the complete input or output for prompt tuning. As illustrated in Figure~\ref{gpc_model}, $P_{t}$ represents the prompt input of the $t$th encoder, while $P_{t}^{}$ represents the prompt output of the $t$th encoder. GPC takes the input and output of the $t$th encoder, $P_{t}$ and $P_{t}^{*}$, to generate the prompt input of the $(t+1)$th encoder $E_t$, $P_{t+1}$. Similarly, $T_{t}$ represents the text input embeddings of the $t$th encoder, while $T_{t}^{}$ represents the text output embeddings of the $t$th encoder, derived from $T_{t}$. Since GPC does not affect text embeddings, $T_{t+1}$ is identical to $T_{t}^{}$. The complete embeddings are updated as follows.
\begin{equation}
 \{P^{*}_{t};T^{*}_t\} =  E_t(\{P_{t};T_{t}\}) 
\end{equation}
 
 The prompt embeddings are updated as below.
\begin{equation}
 P_{t+1} = GPC(P^{*}_{t},P_{t}) 
\end{equation}

\subsection{Inside the GPC}
Figure~\ref{gpc_model} provides a detailed view of the internal structure of GPC. The cell comprises two units: the remembering unit and the forgetting unit. Each unit contains a single-layer feed-forward neural network that is shared across all encoder layers. Similar to ResNet, the remembering unit utilizes the prompt embedding before it is fed to the encoder, in order to emphasize and make use of the original information. In contrast, the forgetting unit operates on the output of the encoder to reduce its influence and prevent extreme outcomes. The cell then combines the outputs of both units to produce the final result. Therefore, between every pair of adjacent encoders, GPC considers both the original and encoded prompt embeddings, enabling the model to retain past information and encouraging it to forget some of the current state, thereby striking a balance between the past and the present. The following equation demonstrates how GPC processes prompt embeddings:

\begin{equation}
 P_{t+1} = \theta(W_{F}P_{t}^{*}+W_{R}P_{t}) 
\end{equation}
where $W_F$ represents the weights of the forgetting unit, while $W_R$ represents the weights of the remembering unit, where $\theta$ is the activation function.

\subsection{Classification Head}

Traditional discrete prompt tuning is mainly used for few-shot or zero-shot tasks, so verbalizers are needed to align with pre-training cloze tasks, in order to narrow the gap between pre-training tasks and downstream tasks, and improve downstream task performance. In contrast, the continuous prompt tuning used in our paper aims to reduce the amount of parameter training and storage during fine-tuning. It generally uses the entire dataset instead of few-shot scenarios, and selecting an appropriate verbalizer manually requires a lot of training resources.

In order to simplify the model and reduce manual intervention as in \cite{lester2021power} and \cite{liu2021gpt}, we replace the original verbalizers with a classification head that receives outputs from the final layer of the model. We use a randomly initialized classification head to predict the label from the output of the [CLS] token.
\section{Experiment}
\begin{table*}[h]
\centering
\begin{tabular}{cccccc}
\toprule
\textbf{Corpus}           & \textbf{Train} & \textbf{Dev} & \textbf{Test} & \textbf{Task} & \textbf{Metrics} \\ 
\midrule
BoolQ & 9427             & 3270           & 3245            & Question Answering            & accuracy             \\
CB                        & 250              & 57             & 250             & Natural Language Inference           & accuracy             \\
COPA                      & 400              & 100            & 500             & Question Answering            & accuracy             \\
RTE                       & 2500             & 278            & 300             & Natural Language Inference           & accuracy             \\
WiC                       & 6000             & 638            & 1400            & World Sense Disambiguation           & accuracy             \\
WSC                       & 554              & 104            & 146             & Co-reference Resolution        & accuracy\\  
\bottomrule
\end{tabular}
\caption{Statistics of SuperGLUE datasets.}
\label{stat-table}
\end{table*}

\begin{table*}[h]
\centering
\renewcommand{\arraystretch}{1.1}
\scalebox{0.85}{
\resizebox{0.9\textwidth}{!}{%
\begin{tabular}{ll clclc l clclc } 
\toprule
\multicolumn{1}{c}{} & \multicolumn{1}{c}{} & \multicolumn{5}{c}{BoolQ}                          & \multicolumn{1}{c}{} & \multicolumn{5}{c}{RTE}             \\ 
\cline{3-7}\cline{9-13}
                     &                      & PT            &  & Prompt-only &  & GPC    &                      & PT            &  & Prompt-only &  & GPC      \\ 
\midrule

BERT           &                      & 67.2 &  & 62.8        &  & \textbf{67.9}          &                      & 53.5          &  & 54.5        &  & \textbf{61.0}  \\
RoBERTa       &                      & 62.3          &  & 62.4        &  & \textbf{63.5} &                      & 58.8 &  & 54.2        &  & \textbf{59.4}           \\ 
\midrule
\midrule

\multicolumn{1}{c}{} & \multicolumn{1}{c}{} & \multicolumn{5}{c}{CB}    & \multicolumn{1}{c}{} & \multicolumn{5}{c}{COPA}               \\ 
\cline{3-7}\cline{9-13}
                     &                      & PT            &  & Prompt-only &  & GPC    &                      & PT            &  & Prompt-only &  & GPC  \\ 
\midrule

BERT  &     & 80.4 &  & 71.4          &  & \textbf{82.1}   
&        & 55.0            &  & 58.0          &  & \textbf{67.0}  \\
RoBERTa   &  & 71.4 &  & 69.6            &  & \textbf{73.2}  &                      & 63.0            &  & 62.0          &  & \textbf{66.0} \\ 

\midrule
\midrule

\multicolumn{1}{c}{}     & \multicolumn{1}{c}{} & \multicolumn{5}{c}{WiC}                           & \multicolumn{1}{c}{} & \multicolumn{5}{c}{WSC}                    \\ 
\cline{3-7}\cline{9-13}
                     &                      & PT            &  & Prompt-only &  & GPC    &                      & PT            &  & Prompt-only &  & GPC    \\ 
\midrule
BERT             &                      & 63.0            &  & 56.4        &  & \textbf{66.9} &                      & 64.4 &  & 64.4          &  & \textbf{65.4}   \\
RoBERTa     &              & 56.9          &  & 54.7        &  & \textbf{69.6} &                      & 64.4 &  & 63.5 &  & \textbf{65.4}          \\
\bottomrule
\end{tabular}
}
}
\caption{Results on SuperGLUE development set. PT: Prompt tuning \cite{lester2021power}; Prompt-only: Prompt tuning with no verbalizer; GPC: Prompt tuning with Global Prompt Cell; \textbf{bold}: the best performance.}
\label{tab:my-table}
\end{table*}

\subsection{Datasets and Metrics}
We evaluate on SuperGLUE~\cite{superglue}. SuperGLUE is a new benchmark styled after GLUE with a new set of more difficult language understanding tasks.
Because our task mainly involves classification tasks, while MultiRC and ReCORD belong to QA tasks, and these two tasks are difficult to handle using prompt learning, as reported by \cite{liu2021pre}. Existing prompt learning methods for these two tasks suffer from significant fluctuations and are difficult to converge. Moreover, the effectiveness of our method in classification tasks has been demonstrated through the other six tasks. Therefore, we selected the other six classification tasks for our experiments. We choose the classification tasks and co-reference resolution tasks, including BoolQ, RTE, CB, COPA, WiC and WSC. Statistics of the selected datasets are in Table~\ref{stat-table}.
We use accuracy as our evaluation metric.

\subsection{Experiment Settings}

\paragraph{Models}
We employ BERT~\cite{devlin2019bert} and RoBERTa~\cite{liu2019roberta} for our model, both of which are based on transformer encoders and are typically used for classification tasks.

\paragraph{Prompt Length}
The length of prompt embeddings has a significant impact on model performance. According to \cite{liu2021p}, different tasks require prompt embeddings of varying lengths to achieve optimal results. Typically, simpler tasks require shorter prompts than more complex ones~\cite{liu2021p}. We experimented with prompt lengths of 16, 32, and 64 and selected the most effective ones among them.

\paragraph{Prompt Initialization}
Prompt embeddings can be initialized in various ways, such as random initialization or utilizing concrete token embedding. Based on the approach described in \cite{gu2021ppt}, we use random initialization in our method.

\paragraph{Training Method}
During the training phase, we freeze the original parameters of the PTMs and only update the Global Prompt Cell, specifically the corresponding weight matrices $W_R$ and $W_F$. Therefore, we only need to store a small number of parameters instead of the parameters of the entire PTM for each downstream task.


\subsection{Main Results}
To assess the effectiveness of Global Prompt Cell, we investigate (i) whether replacing the verbalizer with a classification head causes performance degradation in prompt tuning, and (ii) whether Global Prompt Cell can outperform prompt tuning.
(iii) whether Global Prompt Cell combined with a classification head still yield better results.

We carry out prompt-only experiments, where we substitute the verbalizer with a classification head and refrain from using GPC. Table~\ref{tab:my-table} demonstrates that the performance of prompt-only models decreases, particularly on the CB dataset, with both BERT and RoBERTa models experiencing over 10\% decrease.

For (ii) and (iii), we perform experiments with Global Prompt Cell. As shown in Table~\ref{tab:my-table}, GPC significantly improves the effectiveness of prompt tuning using classification heads, surpassing the performance of vanilla prompt tuning. GPC can achieve over 10\% performance increase compared to the prompt-only model on both WiC and CB datasets. In comparison with the PT (Prompt-Tuning~\cite{lester2021power}) model, which requires multiple experiments to determine the optimal verbalizer, our method still outperforms it on all six tasks.These results demonstrate the efficacy of our method.

\begin{figure*}[htbp] 
\centering 
\includegraphics[width=\textwidth]{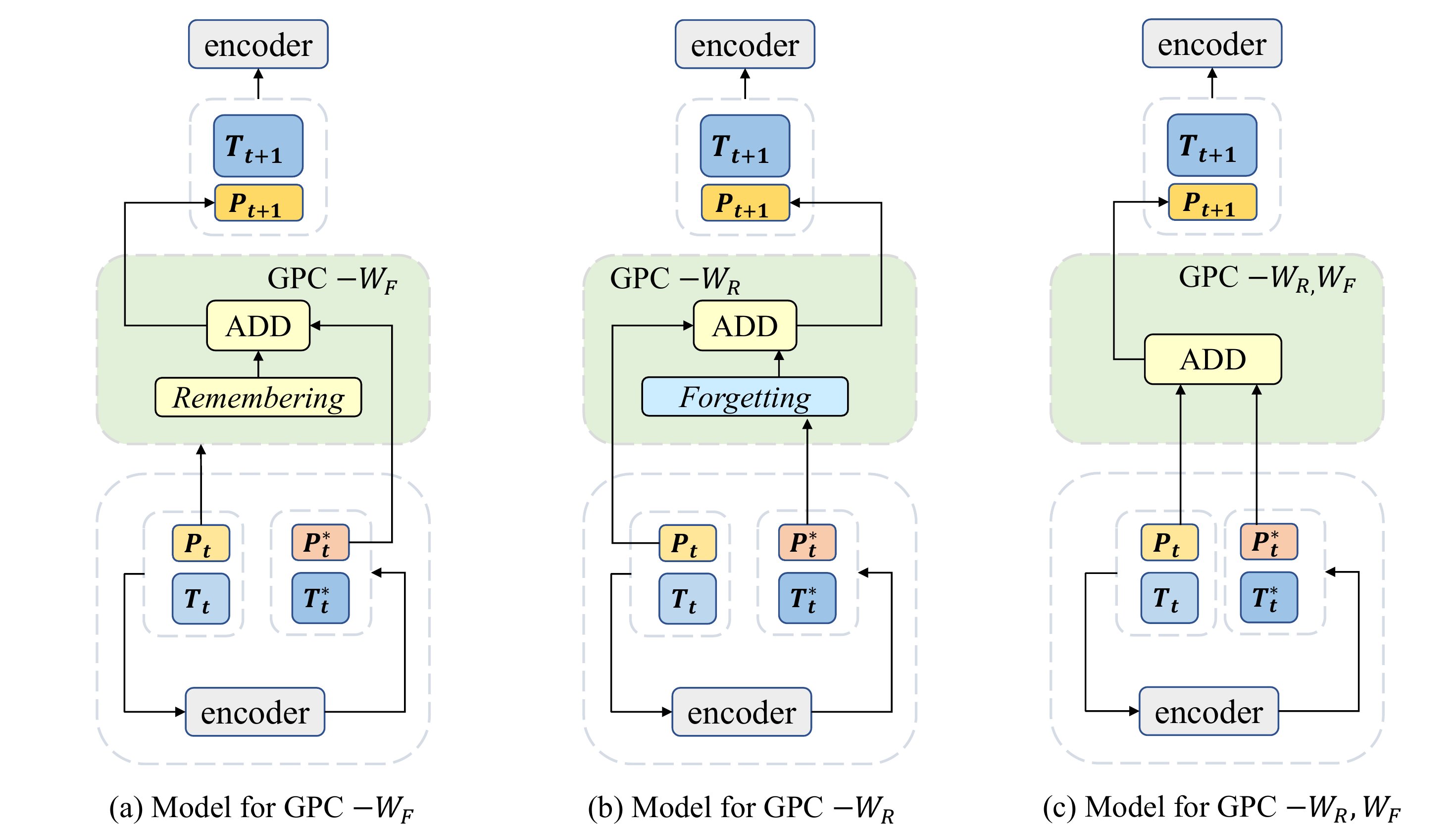} 
\caption{Models in ablation study.} 
\label{ablation-model-fig} 
\end{figure*}
\begin{table}[h]
\centering
\begin{tabular}{ll}
\toprule        Model            &  Updating strategy\\
\midrule
GPC                 & $P_{t+1} = \theta(W_{F}P_{t}^{*}+W_{R}P_{t})$ \\
GPC$-W_{F}$       & $P_{t+1} = \theta(P_{t}^{*}+W_{R}P_{t})$ \\
GPC$-W_{R}$         & $P_{t+1} = \theta(W_{F}P_{t}^{*}+P_{t})$ \\
GPC$-W_{F},W_{R}$ & $P_{t+1} = \theta(P_{t}^{*}+P_{t})$ \\
\bottomrule
\end{tabular}
\caption{Different settings for prompt update. }
\label{ablation-formula}
\end{table}

\begin{table*}[h]
\centering
\begin{tabular}{ll l l l l l l}
\toprule
                            &          & BoolQ & RTE  & CB   & COPA & WiC  & WSC  \\
\midrule       
\multirow{4}{*}{BERT}    & GPC     & \textbf{67.9}  & \textbf{61.0} & \textbf{82.1} & \textbf{67.0} & \textbf{66.9} & \textbf{65.4} \\
\cline{2-8}
                               & GPC$-W_{F}$    & 62.3  & 57.8 & 67.9 & 61.0 & 60.7 & \textbf{65.4}     \\
                           & GPC$-W_{R}$     & 62.6  & 57.8 & 55.4 & 63.0 & 57.5 &  64.4    \\
                              & GPC$-W_{F},W_{R}$ & 62.7  & 58.1 & 75.0 & 63.0 & 57.2 & 63.5     \\
\midrule
\multirow{4}{*}{RoBERTa} & GPC       & \textbf{63.5}  & \textbf{59.4} & \textbf{73.2} & \textbf{66.0} & \textbf{69.6} & \textbf{65.4} \\
\cline{2-8}
                               & GPC$-W_{F}$    & 62.2      & 55.6 & 61.0 & 55.0 & 53.0 & 63.5     \\
                              & GPC$-W_{R}$     & 62.2      & 52.7 & 64.3 & 55.0 & 50.9 & 63.5     \\
                              & GPC$-W_{F},W_{R}$ & 62.3     & 57.0 & 57.1 & 58.0 & 52.2 & 63.5    \\
\bottomrule
\end{tabular}
\caption{Ablation experiment results. GPC: Global Prompt Cell; GPC$-W_{F}$: removing the forgetting unit; GPC$-W_{R}$: removing the remembering unit; GPC-$W_{F},W_{R}$: removing both the forgetting unit and remembering unit. \textbf{Bold}: the best performance.}
\label{ablation-table}
\end{table*}

\section{Ablation Study}

Our GPC module consists of two parts, remembering unit and forgetting unit. The remembering unit receives previous prompt embeddings and selects which parts to retain. The remembering unit can be helpful for retaining certain information that requires long-term memory. The forgetting unit is responsible for determining which parts of the latest prompt embeddings are irrelevant, preventing the addition of unnecessary information.

We design the ablation experiments to explore the effectiveness of the above units. We conducted three groups of ablation experiments, where we added only the memory unit module, only the forgetting unit module, and no module at all. Through this, we aim to demonstrate the impact of memory and forgetting units on the experimental results, as well as prove the necessity and importance of these two modules.

Figure~\ref{ablation-model-fig} shows the architectures of three ablation settings, which respectively remove the forgetting unit, the remembering unit and the both. 
Table~\ref{ablation-formula} shows the formula of the three different updating strategies and the comparison with GPC.

The result in Table~\ref{ablation-table} shows that the contribution of two units varies from task to task, but the combination of the two can reach optimal results for all cases.

In addition to the findings presented in Table 4, further analysis of our ablation experiments revealed that the impact of the remembering and forgetting units on task performance was dependent on the specific task. Specifically, for some tasks, the addition of the memory unit module resulted in a greater improvement in performance compared to the addition of the forgetting unit module, while for other tasks the opposite was true.

The combination of both modules consistently led to optimal results across all tasks. This suggests that while the individual contributions of the remembering and forgetting units may vary depending on the specific task, the integration of both modules is crucial for achieving optimal performance across a range of tasks.


\section{Conclusion}


To conclude, our study has introduced the Global Prompt Cell (GPC) as a novel approach to enhance continuous prompt tuning. By selectively remembering and forgetting prompt embeddings, GPC enables more effective prompt updating, ultimately leading to improved model performance.


Through experiments conducted on the SuperGLUE benchmark, we have shown that our approach can substantially enhance results using prompt tuning. This highlights the potential of GPC to significantly improve the performance of language models on a range of natural language understanding tasks.

Finally, we emphasize that GPC can serve as a portable plug-in module for prompt tuning paradigms, allowing for easy integration with existing models and architectures. We believe that our approach has promising implications for the development of more effective and efficient language models, and we look forward to further exploration and refinement of this method in future research.




\bibliographystyle{splncs04}
\bibliography{main}
\end{document}